\documentclass[11pt,a4paper]{article}
\usepackage[hyperref]{acl2019}
\usepackage{times}
\usepackage{latexsym}
\usepackage{url}
\usepackage{multirow}
\usepackage{graphicx} 
\usepackage{amsmath}  
\usepackage{adjustbox}
\usepackage{algorithm}      
\usepackage{algpseudocode}
\usepackage{bm}             
\usepackage{subfigure}
\usepackage{amsfonts}
\usepackage{makecell}

\aclfinalcopy

\def\aclpaperid{1101} 

\title{GCDT: A Global Context Enhanced Deep Transition Architecture \\ for Sequence Labeling}

\author{
  Yijin Liu\textsuperscript{1}\thanks{\ \ This work was done when Yijin Liu was interning at Pattern Recognition Center, WeChat AI, Tencent Inc, China} \ ,
  Fandong Meng\textsuperscript{2}, 
  Jinchao Zhang\textsuperscript{2}, 
  Jinan Xu\textsuperscript{1}\thanks{ \ \ Jinan Xu is the corresponding author of the paper.} ,
  Yufeng Chen\textsuperscript{1} 
  and Jie Zhou\textsuperscript{2} \\
  \textsuperscript{1}Beijing Jiaotong University, China \\
  \textsuperscript{2}Pattern Recognition Center, WeChat AI, Tencent Inc, China \\
  \texttt{adaxry@gmail.com} \\
  \texttt{\{fandongmeng, dayerzhang, withtomzhou\}@tencent.com} \\
  \texttt{\{jaxu,chenyf\}@bjtu.edu.cn} \\
}

\date{}
\begin{document}
\maketitle
\begin{abstract}
Current state-of-the-art systems for the sequence labeling  tasks are typically based on the family of Recurrent Neural Networks (RNNs). 
However, the shallow connections between consecutive hidden states of RNNs and insufficient modeling of global information restrict the potential performance of those models. In this paper, we try to address these issues, and thus propose a Global Context enhanced Deep Transition architecture for sequence labeling named GCDT.
We deepen the state transition path at each position in a sentence, and further assign every token with a global representation learned from the entire sentence.
Experiments on two standard sequence labeling tasks show that, given only training data and the ubiquitous word embeddings (Glove), our GCDT achieves 91.96 $F_1$ on the CoNLL03 NER task and 95.43 $F_1$ on the CoNLL2000 Chunking task, which outperforms the best reported results under the same settings.
Furthermore, by leveraging BERT as an additional resource, we establish new state-of-the-art results with 93.47 $F_1$ on NER and 97.30 $F_1$ on Chunking \footnote{Code is available at:  https://github.com/Adaxry/GCDT.}. 

\end{abstract}

\section{Introduction}
Sequence labeling tasks, including part-of-speech tagging (POS), syntactic chunking and named entity recognition (NER), are fundamental and challenging problems of Natural Language Processing (NLP).
Recently, neural models have become the de-facto standard for high-performance systems. Among various neural networks for sequence labeling, bi-directional RNNs (BiRNNs), especially BiLSTMs \cite{LSTM} have become a dominant method on multiple benchmark datasets \cite{BLSTM+CRF,char-CNN+BLSTM,char-LSTM+BLSTM+CRF,Peter2017}.

However, there are several natural limitations of the BiLSTMs architecture. For example, at each time step, the BiLSTMs consume an incoming word and construct a new summary of the past subsequence. This procedure should be highly nonlinear, to allow the hidden states to rapidly adapt to the mutable input while still preserving a useful summary of the past \cite{DT_language_model}. While in BiLSTMs, even stacked BiLSTMs, the transition depth between consecutive hidden states are inherently shallow. Moreover, global contextual information, which has been shown highly useful for model sequence \cite{SLSTM}, is insufficiently captured at each token position in BiLSTMs. Subsequently, inadequate representations flow into the final prediction layer, which leads to the restricted performance of BiLSTMs.

In this paper, we present a global context enhanced deep transition architecture to eliminate the mentioned limitations of BiLSTMs. In particular, we base our network on the deep transition (DT) RNN \cite{DT_language_model},  which increases the transition depth between consecutive hidden states for richer representations. Furthermore, we assign each token an additional representation, which is a summation of hidden states of a specific DT over the whole input sentence, namely global contextual embedding. It's beneficial to make more accurate predictions since the combinatorial computing between diverse token embeddings and global contextual embedding can capture useful representations in a way that improves the overall system performance.

We evaluate our GCDT on both CoNLL03 and CoNLL2000. Extensive experiments on two benchmarks suggest that, merely given training data and publicly available word embeddings (Glove), our GCDT surpasses previous state-of-the-art systems on both tasks. Furthermore, by exploiting BERT as an extra resource, we report new state-of-the-art $F_1$ scores with 93.47 on CoNLL03 and 97.30 on CoNLL2000.
The main contributions of this paper can be summarized as follows:
\begin{itemize}
\item We are the first to introduce the deep transition architecture for sequence labeling, and further enhance it with the global contextual representation at the sentence level, named GCDT.

\item GCDT substantially outperforms previous systems on two major tasks of NER and Chunking. Moreover, by leveraging BERT as an extra resource to enhance GCDT, we report new state-of-the-art results on both tasks. 

\item We conduct elaborate investigations of global contextual representation, model complexity and effects of various components in GCDT.
\end{itemize}

\begin{figure*}[t!]
\begin{center}
     \scalebox{0.95}{
       \includegraphics[width=1\textwidth]{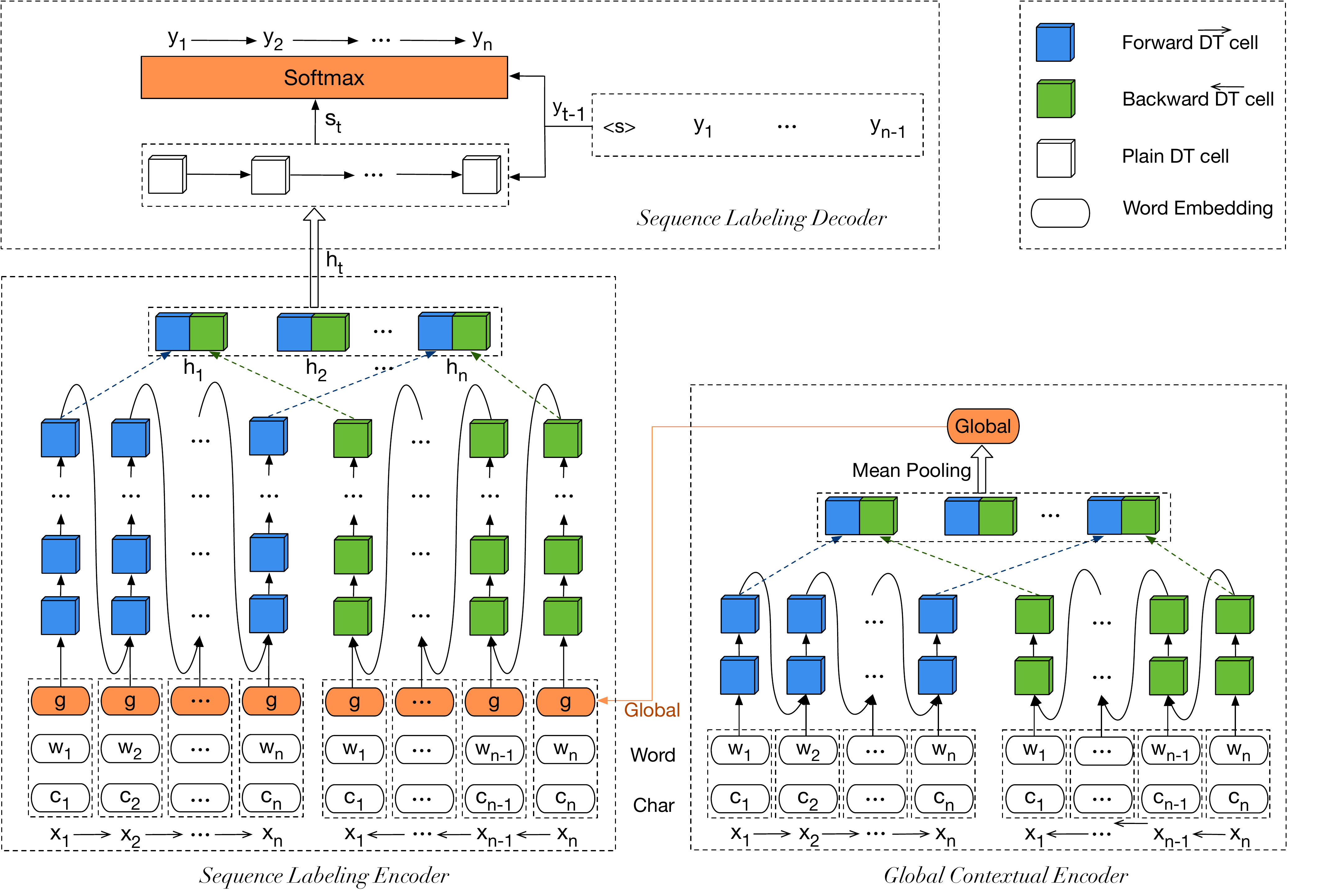}
      }
      \caption{Overview of GCDT. The \emph{global contextual encoder} (on the right) serves as an enhancement of token representation. The \emph{sequence labeling encoder} and \emph{decoder} (on the left) take charge of the task-specific predictions. } \label{overview}  \vspace{-2pt}
 \end{center} \vspace{-2pt}
\end{figure*}

\section{Background}
Given a sequence of $X = \{x_1, x_2, \cdots, x_N \}$ with $N$ tokens and its corresponding linguistic labels $Y = \{y_1, y_2, \cdots, y_N \}$ with the equal length, the sequence labeling tasks aim to learn a parameterized mapping function $f_{\theta} : X
\rightarrow Y $ from input tokens to task-specific labels.

Typically, the input sentence is firstly encoded into a sequence of distributed representations $\mathbf{X} = \{\mathbf{x}_1, \mathbf{x}_2, \cdots, \mathbf{x}_N\}$ by character-aware and pre-trained word embeddings.
The majority of high-performance models use bidirectional RNNs, BiLSTMs in particular, to encode the token embeddings $\mathbf{X}$ into context-sensitive representations for the final prediction. 

Additionally, it's beneficial to model and predict labels jointly, thus a subsequent conditional random field (CRF \citealp{CRF}) is commonly utilized as a decoder layer. At the training stage, those models maximize the log probability of the correct sequence of tags as follows: 
\begin{align}
\log (p (\mathbf{y}|\mathbf{X}) ) = s(\mathbf{X},\mathbf{y}) - \log (\sum_{\widetilde{\mathbf{y}} \in  \mathbf{Y}_x } e ^ { s (\mathbf{X}, \widetilde{\mathbf{y}}) } )
\end{align}
where $s(\cdot)$ is the score function and $\mathbf{Y}_x$ is the set of all possible sequence of tags. Typically, the Viterbi algorithm \cite{viterbi} is utilized to search the label sequences with maximum score when decoding:
\begin{align}
y^{*} = \mathop{\arg\max}_{\widetilde{y} \in Y_x } s (x, \widetilde{y} )
\end{align}

\section{GCDT}

\subsection{Overview}
In this section, we start with a brief overview of our presented GCDT and then proceed to structure the following sections with more details about each submodule. As shown in Figure \ref{overview}, there are three deep transition modules in our model, namely \emph{global contextual encoder}, \emph{sequence labeling encoder} and \emph{decoder} accordingly. 
\paragraph{Token Representation}
Given a sentence $X = \{x_1, x_2, ..., X_N\}$ with $N$ tokens,
our model first captures each token representation $\mathbf{x}_t$ by concatenating three primary embeddings: 
\begin{align}
& \mathbf{x}_t = [\mathbf{c}_t; \mathbf{w}_t; \mathbf{g}] \label{token_embedding} 
\end{align}

\begin{enumerate}
\item Character level word embedding $\mathbf{c}_t$ is acquired from Convolutional Neural Network. (CNN) \cite{first_CNN}
\item Pre-trained word embedding $\mathbf{w}_t$ is obtained from the lookup table initialized by Glove\footnote{https://nlp.stanford.edu/projects/glove/}.
\item Global contextual embedding $\mathbf{g}$ is extracted from bidirectional DT, and more details will be described in the following paragraphs.
\end{enumerate}
The global embedding $\mathbf{g}$ is computed by mean pooling over all hidden states $\{\mathbf{h}_1^g, \mathbf{h}_2^g, \cdots, \mathbf{h}_N^g\}$ of \emph{global contextual encoder} (right part in Figure \ref{overview}). For simplicity, we can take ``DT" as a reinforced Gated Recurrent Unit (GRU \citealp{GRU}), and more details about DT will be described in the next section. Thus $\mathbf{g}$ is computed as follows:
\begin{align}
& \mathbf{g} = \frac{1}{N}\sum_{t=1}^n{\mathbf{h}_{t}^g} \\
& \mathbf{h}_{t}^g = [\overrightarrow{\mathbf{h}}_{t}^g; \overleftarrow{\mathbf{h}}_{t}^g] \\
& \overrightarrow{\mathbf{h}}_{t}^g = \overrightarrow{\mathbf{DT}}_{g} (\mathbf{c_t}, \mathbf{w_t}; \mathbf{\theta}_{\overrightarrow{{DT}}_{g}}) \\
& \overleftarrow{\mathbf{h}}_{t}^g = \overleftarrow{\mathbf{DT}}_{g} (\mathbf{c_t}, \mathbf{w_t}; \mathbf{\theta}_{\overleftarrow{{DT}}_{g}})
\end{align}

\paragraph{Sequence Labeling Encoder}
Subsequently, the concatenated token embeddings $\mathbf{x}_t$ (Eq. \ref{token_embedding}) is fed into the \emph{sequence labeling encoder} (bottom left part in Figure \ref{overview}).
\begin{align}
& \mathbf{h}_t = [\overrightarrow{\mathbf{h}_t}; \overleftarrow{\mathbf{h}_t}] \\
& \overrightarrow{\mathbf{h}_t} = \overrightarrow{\mathbf{DT}}_{en}(\mathbf{x}_t, \overrightarrow{\mathbf{h}}_{t-1}; \mathbf{\theta}_{\overrightarrow{{DT}}_{en}}) \\
& \overleftarrow{\mathbf{h}_t} = \overleftarrow{\mathbf{DT}}_{en}(\mathbf{x}_t, \overleftarrow{\mathbf{h}}_{t-1}; \mathbf{\theta}_{\overleftarrow{{DT}}_{en}})
\end{align}

\paragraph{Sequence Labeling Decoder} 
\ Considering the $t$-th word in this sentence, the output of \emph{sequence labeling encoder} $\mathbf{h_t}$ along with the past label embedding $\mathbf{y}_{t-1}$ are fed into the \emph{decoder} (top left part in Figure \ref{overview}). Subsequently, 
the output of \emph{decoder} $\mathbf{s}_t$ is transformed into $\mathbf{l}_t$ for the final softmax over the tag vocabulary. Formally, the label of word $\mathbf{x}_t$ is predicted as the probabilistic equation (Eq. \ref{final_predict})
\begin{align}
& \mathbf{s}_t = \mathbf{DT}_{de}(\mathbf{h}_t, \mathbf{y}_{t-1}; \mathbf{\theta}_{DT_{de}}) \\
& \mathbf{l}_t = \mathbf{s}_t \mathbf{W}_l + \mathbf{b}_l \\
& P (y_t = j | \mathbf{x}) = softmax(\mathbf{l}_t) [j] \label{final_predict} 
\end{align}
As we can see from the above procedures and Figure \ref{overview}, our GCDT firstly encodes the global contextual representation along the sequential axis by DT, which is utilized to enrich token representations.
At each time step, we encode the past label information jointly using the \emph{sequence labeling decoder} instead of resorting to CRF.
Additionally, we employ beam search algorithm to infer the most probable sequence of labels when testing.

\subsection{Deep Transition RNN}
Deep transition RNNs extend conventional RNNs by increasing the transition depth of consecutive hidden states. Previous studies have shown the superiority of this architecture on both language modeling \cite{DT_language_model} and machine translation \cite{DT_machine_translation,DTMT}. Particularly, \citeauthor{DTMT} \shortcite{DTMT} propose to maintain a linear transformation path throughout the deep transition procedure with a linear gate to enhance the transition structure.

Following \citeauthor{DTMT} \shortcite{DTMT}, the deep transition block in our hierarchical model is composed of two key components, namely Linear Transformation enhanced GRU (L-GRU) and Transition GRU (T-GRU). At each time step, L-GRU first encodes each token with an additional linear transformation of the input embedding, then the hidden state of L-GRU is passed into a chain of T-GRU connected merely by hidden states. Afterwards, the output ``state" of the last T-GRU for the current time step is carried over as ``state" input of the first L-GRU for the next time step. Formally, in a unidirectional network with transition number of $L$, the hidden state of the $t$-th token in a sentence is computed as:
\begin{align}
& \mathbf{h}_{i}^0 = \mathbf{L}\textrm{-}\mathbf{GRU}(\mathbf{x}_i, \mathbf{h}_{i-1}^L)  \\
& \mathbf{h}_{i}^j = \mathbf{T}\textrm{-}\mathbf{GRU}^{j}( \mathbf{h}_{i}^{j-1})  \ \ \ \ \ \ \   1 \leq j \leq L  
\end{align}

\paragraph{Linear Transformation Enhanced GRU}
L-GRU extends the conventional GRU by an additional linear transformation of the input token embeddings. At time step $t$, the hidden state of L-GRU is computed as follows:
\begin{align}
& \mathbf{h}_t = (1 - \mathbf{z}_t) \odot \mathbf{h}_{t-1} + \mathbf{z}_{ t } \odot \widetilde {\mathbf{h}}_t  \ \ \ \ \ \ \ \ \ \ \ \ \ \\
& \begin{aligned}
   \widetilde{\mathbf{h}_t} = \tanh ( & \mathbf{W}_{xh} \mathbf{x}_{t} + \mathbf{r}_{t}  \odot (\mathbf{W}_{hh} \mathbf{h}_{t-1})) \\ 
    & + \mathbf{l}_{t} \odot \mathbf{W}_x \mathbf{x}_t
  \end{aligned}
  \label{L-GRU_h_t}
\end{align}
where $\mathbf{W}_{xh}$ and $\mathbf{W}_{hh}$ are parameter matrices, and reset gate $\mathbf{r}_t$ and update gate $\mathbf{z}_t$ are same as GRU:
\begin{align}
& \mathbf{r}_t = \sigma (\mathbf{W}_{xr} x_t + \mathbf{W}_{hr} \mathbf{h}_{t-1} ) \\
& \mathbf{z}_t = \sigma (\mathbf{W}_{xz} \mathbf{x}_t + \mathbf{W}_{hz} \mathbf{h}_{t-1}) 
\end{align}
The linear transformation $\mathbf{W}_x \mathbf{x}_t$ in candidate hidden state $\widetilde{\mathbf{h}}_t$ (Eq. \ref{L-GRU_h_t}) is regulated by the linear gate $\mathbf{l}_t$, which is computed as follows:
\begin{align}
\mathbf{l}_t = \sigma (\mathbf{W}_{xl} \mathbf{x}_t + \mathbf{W}_{hl} \mathbf{h}_{t-1})
\end{align}
\paragraph{Transition GRU} T-GRU is a special case of conventional GRU, which only takes hidden states from the adjacent lower layer as inputs. At time step $t$ at transition depth $l$, the hidden state of T-GRU is computed as follows:
\begin{align}
  \mathbf{h}^{l}_{t} = (1 - \mathbf{z}^{l}_{t}) \odot \mathbf{h}^{l - 1}_{t} + \mathbf{z}^{l}_{t} \odot \widetilde {\mathbf{h}_{t}}^l \\
  \widetilde{\mathbf{h}_{t}}^l = \tanh (\mathbf{r}_{t}^l \odot (\mathbf{W}^{l}_{h} \mathbf{h}_{t}^{l-1})) \ \ \ \ \ \ \  
\end{align}

Reset gate $r_t$ and update gate $z_t$ also only take hidden states as input, which are computed as:
\begin{align}
\mathbf{r}^l = \sigma(\mathbf{W}^{l}_{r} \mathbf{h}^{l-1}) \\
\mathbf{z}^t = \sigma(\mathbf{W}^{l}_{z} \mathbf{h}^{l-1})
\end{align}

As indicated above, at each time step of our deep transition block, there is a L-GRU in the bottom and several T-GRUs on the top of L-GRU.

\subsection{Local Word Representation}
\paragraph{Charater-aware word embeddings} 
It has been demonstrated that character level information (such as capitalization, prefix and suffix) \cite{collobert2011,first_CNN} is crucial for sequence labeling tasks. In our GCDT, the character sets consist of all unique characters in datasets besides the special symbol ``PAD" and ``UNK". We use one layer of CNN followed by max pooling to generate character-aware word embeddings.

\paragraph{Pre-trained word embeddings} 
The pre-trained word embeddings have been indicated as a standard component of neural network architectures for various NLP tasks. Since the capitalization feature of words is crucial for sequence labeling tasks \cite{collobert2011}, we adopt word embeddings trained in the case sensitive schema. 

Both the character-aware and pre-trained word embeddings are context-insensitive, which are called local word representations compared with global contextual embedding in the next section.

\subsection{Global Contextual Embedding}
We adopt an independent deep transition RNN named \emph{global contextual encoder} (right part in Figure \ref{overview}) to capture global features. In particular, we transform the hidden states of \emph{global contextual encoder} into a fixed-size vector with various strategies, such as mean pooling, max pooling and self-attention mechanism \cite{attention}. According to the preliminary experiments, we choose mean pooling strategy considering the balance between effect and efficiency.

In conventional BiRNNs, the global contextual feature is insufficiently modeled at each position, as the nature of recurrent architecture makes RNN partial to the most recent input token. While our context-aware representation is incorporated with local word embeddings directly, which assists in capturing useful representations through combinatorial computing between diverse local word embeddings and the global contextual embedding. We further investigate the effects on positions where the global embedding is used. (Section \ref{section_global_position})

\section{Experiments}

\subsection{Datasets and Metric}

\paragraph{NER}
The CoNLL03 NER task \cite{CoNLL2003} is tagged with four linguistic entity types (PER, LOC, ORG, MISC). Standard data includes train, development and test sets. 

\paragraph{Chunking}
The CoNLL2000 Chunking task \cite{CoNLL2000} defines 11 syntactic chunk types (NP, VP, PP, {\em etc.}). Standard data includes train and test sets.

\paragraph{Metric}
We adopt the BIOES tagging scheme for both tasks instead of the standard BIO2, since previous studies have highlighted meaningful improvements with this scheme \cite{BIES}. We take the official conlleval \footnote{https://www.clips.uantwerpen.be/conll2000/chunking/\\conlleval.txt} as the token-level $F_1$  metric. Since the data size if relatively small, we train each final model for 5 times with different parameter initialization and report the mean and standard deviation $F_1$ value. 

\subsection{Implementation Details}
All trainable parameters in our model are initialized by the method described by \citeauthor{Xavier} \shortcite{Xavier}. We apply dropout \cite{dropout} to embeddings and hidden states with a rate of 0.5 and 0.3 respectively. All models are optimized by the Adam optimizer \cite{Adam} with gradient clipping of 5 \cite{gradient_clip}. 
The initial learning rate $\alpha$ is set to 0.008, and decrease with the growth of training steps. We monitor the training process on the development set and report the final result on the test set. 
One layer CNN with a filter of size 3 is utilized to generate 128-dimension word embeddings by max pooling. The cased, 300d Glove is adapted to initialize word embeddings, which is frozen in all models. In the auxiliary experiments, the output hidden states of BERT are taken as additional word embeddings and kept fixed all the time.

Empirically, We assign the following hyper-parameters with default values except 
mentioned later.
We set batch size to 4096 at the token level, transition number to 4, hidden size of \emph{sequence labeling encoder} and \emph{decoder} to 256, hidden size of \emph{global contextual encoder} to 128. 

\begin{table}[t!]
\begin{center}
\scalebox{0.9}{
\begin{tabular}{l|c}
\hline \textbf{Models} & \textbf{ $F_1$ } \\ \hline
\cite{collobert2011}* & 89.59 \\
\cite{BLSTM+CRF}* & 90.10 \\
\cite{Lex_emb}* & 90.90 \\
\cite{char-LSTM+BLSTM+CRF} & 90.94 \\
\cite{joint_Yang2016}* & 90.94\\
\cite{joint_Luo2015}* & 91.20 \\
\cite{char-CNN+BLSTM+CRF} & 91.21 \\
\cite{Transfer_Yang2017}*$\dagger$ & 91.26 \\
\cite{SLSTM} & 91.57 \\ 
\cite{reranking} & 91.62 \\
\cite{char-CNN+BLSTM}*$\dagger$ & 91.62 $\pm$ 0.33 \\
\cite{deep_char_cnn} & 91.64 $\pm$ 0.17 \\
\hline
\textbf{GCDT} & \textbf{91.96 $\pm$ 0.04} \\
\textbf{GCDT + $BERT_{LARGE}$} & \textbf{93.47 $\pm$ 0.03} \\
\hline
\end{tabular}}
\end{center}
\caption{ $F_1$ scores on CoNLL03. $\dagger$ refers to models trained on both training and development set. * refers to adopting external task-specific resources.}
\label{ner_result}
\end{table}
\begin{table}[t!]
\begin{center}
\scalebox{0.9}{
\begin{tabular}{l|c}
\hline \textbf{Models} & \textbf{ $F_1$ } \\ \hline
\cite{collobert2011}* & 94.32 \\
\cite{BLSTM+CRF}* & 94.46 \\
\cite{Transfer_Yang2017} & 94.66 \\
\cite{three_chunking} & 94.72 \\
\cite{one_chunking} & 95.02 \\
\cite{two_chunking} & 95.28 \\
\cite{deep_char_cnn} & 95.29 $\pm$ 0.08 \\
\hline
\textbf{GCDT} & \textbf{95.43 $\pm$ 0.06} \\
\textbf{GCDT + $BERT_{LARGE}$} & \textbf{97.30 $\pm$ 0.03} \\
\hline
\end{tabular}}
\end{center}
\caption{ $F_1$ scores on CoNLL2000 Chunking task. * refers to adopting external task-specific resources (like Gazetteers or annotated data).}
\label{chunking_result}
\end{table}

\subsection{Main Results}
The main results of our GCDT on the CoNLL03 and CoNLL2000 are illustrated in Table \ref{ner_result} and Table \ref{chunking_result} respectively. 
Given only standard training data and publicly available word embeddings, our GCDT achieves state-of-the-art results on both tasks. It should be noted that some results on NER are not comparable to ours directly, as their final models are trained on both training and development data \footnote{We achieve $F_1$ score of 92.18 when training on both training and development data without extra resources.}. More notably, our GCDT surpasses the models that exploit additional task-specific resources or annotated corpora \cite{joint_Luo2015,Transfer_Yang2017,char-CNN+BLSTM}.

Additionally, we conduct experiments by leveraging the well-known BERT as an external resource for relatively fair comparison with models that utilize external language models trained on massive corpora.
Especially, \citeauthor{LM_no_annotation} \shortcite{LM_no_annotation} and   \citeauthor{LM_liu2017} \shortcite{LM_liu2017} build task-specific language models only on supervised data.
Table \ref{ner_result_LM} and Table \ref{chunking_result_LM} show that our GCDT outperforms previous state-of-the-art results substantially at 93.47 (+0.38) on NER and 97.30 (+0.30) on Chunking when contrasted with a collection of highly competitive baselines.

\begin{table}[t!]
\begin{center}
\scalebox{0.9}{
\begin{tabular}{l|c}
\hline \textbf{Models} & \textbf{ $F_1$ } \\ \hline
\cite{LM_no_annotation} &  86.26 \\
\cite{LM_liu2017} & 91.71 $\pm$ 0.10 \\
\cite{Peter2017}$\dagger$ & 91.93 $\pm$ 0.19 \\
\cite{EMLo} & 92.20 \\
\cite{EMLo+CV} & 92.61 \\
\shortcite{BERT} $BERT_{BASE}$ & 92.40 \\
\shortcite{BERT} $BERT_{LARGE}$ & 92.80 \\
\cite{contextual_emb}$\dagger$ & 93.09 \\
\hline
\textbf{GCDT + $BERT_{LARGE}$} & \textbf{93.47 $\pm$ 0.03} \\
\hline
\end{tabular}}
\end{center}
\caption{ $F_1$ scores on the CoNL03 NER task by leveraging language model, $\dagger$ refers to models trained on both training and development data. We establish new state-of-the-art result on this task.}
\label{ner_result_LM}
\end{table}
\begin{table}[t!]
\begin{center}
\scalebox{0.9}{
\begin{tabular}{l|c}
\hline \textbf{Models} & \textbf{ $F_1$ } \\ \hline
\cite{LM_no_annotation} & 93.88 \\
\cite{LM_liu2017} & 95.96 $\pm$ 0.08 \\
\cite{Peter2017} & 96.37 $\pm$ 0.05 \\
\cite{contextual_emb} & 96.72 $\pm$ 0.05 \\
\cite{EMLo+CV} & 97.00 \\
\hline
\textbf{GCDT + $BERT_{LARGE}$} & \textbf{97.30 $\pm$ 0.03} \\
\hline
\end{tabular}}
\end{center}
\caption{ $F_1$ scores on the CoNLL2000 Chunking task by leveraging language model. We establish new state-of-the-art result on this task.}
\label{chunking_result_LM}
\end{table}

\section{Analysis}
We choose the CoNLL03 NER task as example to elucidate the properties of our GCDT and conduct several additional experiments.

\subsection{Where to Use the Global Representation?} \label{section_global_position}
In this experiment, we investigate the effects of locations on the global contextual embedding in our hierarchical model. In particular, we use the global embedding $\mathbf{g}$ to augment:

\begin{itemize}
\item \textbf{input of final softmax layer }; \\ $\mathbf{x}_{k}^{softmax} = [\mathbf{h}_k^{decoder} ; \mathbf{y}_{k-1} ; \mathbf{g}]$
\item \textbf{input of sequence labeling decoder}; \\ 
$\mathbf{x}_{k}^{decoder} = [\mathbf{h}_k^{encoder} ; \mathbf{y}_{k-1}; \mathbf{g}]$
\item \textbf{input of sequence labeling encoder}; \\
$\mathbf{x}_{k}^{encoder} = [\mathbf{w}_k ; \mathbf{c}_k; \mathbf{g}]$
\end{itemize}

\begin{table}[t!]
\begin{center}
\scalebox{0.9}{
\begin{tabular}{c|l|c}
\hline \textbf{\#} & \textbf{Use global embedding at} & \textbf{ $F_1$ } \\ \hline
0 & None & 91.60 \\
1 & Input of final softmax & 91.48 \\
2 & Input of sequence labeling decoder & 91.45 \\
3 & Input of sequence labeling encoder & \textbf{91.96} \\
\hline
\end{tabular}}
\end{center}
\caption{Comparison of CoNLL03 test $F_1$  when the global contextual embedding is used at different layers.}
\label{global_emb_postion}
\end{table}

Table \ref{global_emb_postion} shows that the global embedding $\mathbf{g}$ improves performance when utilized at the relative low layer (row 3) , while $\mathbf{g}$ may do harm to performances when adapted at the higher layers (row 0 vs. row 1 \& 2). In the last option, $\mathbf{g}$ is incorporated to enhance the input token representation for \emph{sequence labeling encoder}, the combinatorial computing between the multi-granular local word embeddings ($\mathbf{w}_k$ and $\mathbf{c}_k$) and global embedding $\mathbf{g}$ can capture more specific and richer representations for the prediction of each token, and thus improves overall system performance. While the other two options (row 1, 2) concatenate the highly abstract $\mathbf{g}$ with hidden states ($\mathbf{h}_k^{encoder}$ or $\mathbf{h}_k^{decoder}$) from the higher layers, which may bring noise to token representation due to the similar feature spaces and thus hurt task-specific predictions.

\subsection{Comparing with Stacked RNNs}
Although our proposed GCDT bears some resemblance to the conventional stacked RNNs, they are very different from each other. Firstly, although the stacked RNNs can process very deep architectures, the transition depth between consecutive hidden states in the token level is still shallow.

Secondly, in the stacked RNNs, the hidden states along the sequential axis are simply fed into the corresponding positions of the higher layers, namely only position-aware features are transmitted in the deep architecture. While in GCDT, the internal states in all token position of the \emph{global contextual encoder} are transformed into a fixed-size vector.
This contextual-aware representation provides more general and informative features of the entire sentence compared with stacked RNNs.

To obtain rigorous comparisons, we stack two layers of deep transition RNNs instead of conventional RNNs with similar parameter numbers of GCDT. 
According to the results in Table \ref{stack-RNN},
the stacked-DT improves the performance of the original DT slightly, while there is still a large margin between GCDT and the stacked-DT. 
As we can see, our GCDT achieves a much better performance than stacked-DT with a smaller parameter size, which further verifies that our GCDT can effectively leverage global information to learn more useful representations for sequence labeling tasks. 

\begin{table}[t!]
\begin{center}
\scalebox{0.9}{
\begin{tabular}{c|c|c}
\hline \textbf{ Model } & \textbf{ \# Parameters } & \textbf{ $F_1$ } \\ \hline
DT & 5.6M & 91.60 \\
stacked-DT & 8.4M & 91.61 \\
GCDT & 7.4M & \textbf{91.96} \\
\hline
\end{tabular}}
\end{center}
\caption{Comparison of CoNLL03 test $F_1$ between stacked RNNs and GCDT.}
\label{stack-RNN}
\end{table}

\subsection{Ablation Experiments} 
We conduct ablation experiments to investigate the impacts of various components in GCDT. More specifically, we remove one kind of token embedding from char-aware, pre-trained and global embeddings for \emph{sequence labeling encoder} each time, and utilize DT or conventional GRU with similar model sizes \footnote{To avoid the effect of various model size, we fine tuning hidden size of each model, and more details in Section \ref{section_model_complexity}}.
Results of different combinations are presented in Table \ref{quantiative investigation}.

Given the same input embeddings, DT surpasses the conventional GRU substantially in most cases, which further demonstrates the superiority of DT in sequence labeling tasks.
Our observations on character-level and pre-trained word embeddings suggest that they have a significant impact on highly competitive results (row 1 \& 3 vs. row 5), which is consistent with previous work \cite{first_CNN,char-LSTM+BLSTM+CRF}. 
Furthermore, the global contextual embedding substantially improves the performances on both DT and GRU based models (row 6 \& 7 vs. row 4 \& 5).

\begin{table}[t!]
\begin{center}
\scalebox{0.9}{
\begin{tabular}{c|c|c|c}
\hline  \textbf{ \# } & \textbf{ Embeddings } & \textbf{ RNN } & \textbf{ $F_1$ } \\ \hline
0 & No char & GRU & 91.14 \\ 
1 & No char & DT & 90.94 \\
2 & No Glove & GRU & 87.23 \\
3 & No Glove & DT & 88.59 \\
4 & No global & GRU & 91.32 \\
5 & No global & DT & 91.60 \\
6 & All & GRU & 91.42 \\
7 & All & DT & \textbf{91.96} \\
\hline
\end{tabular}}
\end{center}
\caption{Ablation experiments on the CoNLL03 to investigate the impacts of various components, where ``char" indicates character-aware word embeddings, ``Glove" indicates pre-trained word embeddings, and ``global" indicates global contextual embedding.}
\label{quantiative investigation}
\end{table}

\begin{table}[t!]
\begin{center}
\scalebox{0.9}{
\begin{tabular}{c|c|c|c}
\hline \multicolumn{3}{c|}{\textbf{BERT}} & \multirow{2}*{\textbf{ $F_1$ }} \\
\cline{0-2}
\textbf{ Type } & \textbf{ Layer }  & \textbf{ Pooling } & ~ \\ \hline
\multirow{6}*{BASE} & \multirow{3}*{6} & first & 92.70 \\ 
~ & ~ & max & 92.88 \\ 
~ & ~ & mean & 92.99 \\ \cline{2-4} 
~ & \multirow{3}*{12} & first & 92.89 \\ 
~ & ~ & max & 92.74 \\  
~ & ~ & mean & 92.92 \\ \cline{0-3}
\multirow{9}*{LARGE} & \multirow{3}*{12} & first & 92.88 \\ 
~ & ~ & max & 93.23 \\ 
~ & ~ & mean & 93.36 \\ \cline{2-4}
~ & \multirow{3}*{18} & first & 93.18 \\ 
~ & ~ & max & 93.07 \\ 
~ & ~ & mean & \textbf{93.47} \\ \cline{2-4} 
~ & \multirow{3}*{24} & first & 92.57 \\
~ & ~ & max & 92.60 \\
~ & ~ & mean & 92.83 \\ 
\hline
\end{tabular}}
\end{center}
\caption{ Comparison of CoNLL03 $F_1$ scores when various types, layers and pooling strategies of BERT are employed. ``first" indicates the first sub-word embedding, ``mean" and ``max" refer to mean and max pooling correspondingly.}
\label{effect_BERT}
\end{table}

\begin{table*}[t!]
\begin{center}
\scalebox{0.9}{
\begin{tabular}{c|c|c|c|c}
\hline \textbf{ \# } & \textbf{Global Contextual Encoder} & \textbf{Sequence Labeling Module} & \textbf{\# Parameters} & \textbf{$F_1$} \\ \hline
0 & GRU-384 & GRU-384 & 7.8M & 91.42 \\
1 & GRU-384 & DT4-256 & 9.5M & 91.53 \\
2 & GRU-512 & DT4-256 & 11.2M & 91.49 \\
3 & DT2-128 & GRU-384 & 5.7M & 91.45 \\
4 & DT2-128 & DT4-256 & 7.2M & 91.72 \\
5 & DT4-128 & DT4-256 & 7.4M & \textbf{91.96} \\
\hline
\end{tabular}}
\end{center}
\caption{ $F_1$ scores on the CoNLL03 and parameter sizes of various models, where ``GRU-384" indicates the conventional GRU with hidden size of 384, while ``DT2-128" refers to deep transition RNN with transition number of 2 and hidden size of 128, similarly for ``DT4-256".}
\label{model_complexity}
\end{table*}

\subsection{Effect of BERT}
WordPiece is adopted to tokenize sequence in BERT, which may cut a word into pieces, such as converting ``Johanson" into ``Johan \#\#son". Therefore, additional efforts should be taken to maintain alignments between input tokens and their corresponding labels.
Three strategies are conducted to obtain the exclusive BERT embedding of each token in a sequence. Firstly, we take the first sub-word as the whole word embedding after tokenization, which is employed in the original paper of BERT \cite{BERT}. Mean and max poolings are used as the latter two strategies. Results of various combinations of BERT type, layer and pooling strategy are illustrated in Table \ref{effect_BERT}.

It's reasonable that BERT trained on large model surpasses the smaller one in most cases due to the larger model capacity and richer contextual representation. For the pooling strategy, ``mean" is considered to capture more comprehensive representations of rare words than ``first" and ``max", thus better average performances. Additionally, we hypothesize that the higher layers in BERT encode more abstract and semantic features, while the lower ones prefer general and syntax information, which is more helpful for our NER and Chunking tasks. These hypotheses are consistent with results emerged in Table \ref{effect_BERT}.

\subsection{Model Complexity} \label{section_model_complexity}
One way of measuring the complexity of a neural model is through the total number of trainable parameters. In GCDT, the \emph{global contextual encoder} increases parameter numbers of the \emph{sequence labeling encoder} due to the enlargement of input dimensions, thus we run additional experiments to verify whether the increment of parameters has a great affection on performances. Empirically, we replace DT with conventional GRU in the \emph{global contextual encoder} and sequence labeling module (both \emph{encoder} and \emph{decoder}) respectively. Results of various combinations are shown in Table \ref{model_complexity}.

Observations on parameter numbers show that DT outperforms GRU substantially, with a smaller size (row 4 \& 5 vs. row 0). From the perspective of \emph{global contextual encoder}, DT gives slightly better result compared with GRU (row 3 vs. row 0). We observe similar results in the sequence labeling module (row 1 \& 2 vs. row 0). Intuitively, it should further improve performance when utilizing DT in both modules, which is consistent with the observations in Table \ref{model_complexity} (row 4 \& 5 vs. row 0).

\section{Related Work}

\paragraph{Neural Sequence Labeling}
\citeauthor{collobert2011} \shortcite{collobert2011} propose a seminal neural architecture for sequence labeling, which learns useful representation from pre-trained word embeddings  instead of hand-crafted features.
\citeauthor{BLSTM+CRF} \shortcite{BLSTM+CRF} develop the outstanding BiLSTMs-CRF architecture, which is improved by incorporating character-level LSTM \cite{char-LSTM+BLSTM+CRF}, GRU \cite{joint_Yang2016}, CNN \cite{first_CNN, deep_char_cnn}, IntNet \cite{deep_char_cnn}. 
The shallow connections between consecutive hidden states in those models inspire us to deepen the transition path for richer representation.

More recently, there has been a growing body of work exploring to leverage language model trained on massive corpora in both character level \cite{Peter2017,EMLo,contextual_emb} and token level \cite{BERT}. Inspired by the effectiveness of language model embeddings, we conduct auxiliary experiments by leveraging the well-known BERT as an additional feature. 

\paragraph{Exploit Global Information}
\citeauthor{global_information} \shortcite{global_information} explore the usage of global feature in the whole document by the co-occurrence of each token, which is fed into a maximum entropy classifier.
With the widespread application of distributed word representations \cite{word2vec} and neural networks \cite{collobert2011,BLSTM+CRF} in sequence labeling tasks, the global information is encoded into hidden states of BiRNNs. Specially, \citeauthor{reranking} \shortcite{reranking} leverage global sentence patterns for NER reranking. Inspired by the global sentence-level representation in S-LSTM \cite{SLSTM}, we propose a more concise approach to capture global information,  which has been demonstrated more effective on sequence lableing tasks.

\paragraph{Deep Transition RNN}
Deep transition recurrent architecture extends conventional RNNs by increasing the transition depth between consecutive hidden states. Previous studies have shown the superiority of this architecture on both language model \cite{DT_language_model} and machine translation \cite{DT_machine_translation,DTMT}. We follow the deep transition architecture in \cite{DTMT}, and extend it into a hierarchical model with the global contextual representation at the sentence level for sequence labeling tasks.

\section{Conclusion}
We propose a novel hierarchical neural model for sequence labeling tasks (GCDT), which is based on the deep transition architecture and motivated by global contextual representation at the sentence level. Empirical studies on two standard datasets suggest that GCDT outperforms previous state-of-the-art systems substantially on both CoNLL03 NER task and CoNLL2000 Chunking task without additional corpora or task-specific resources. Furthermore, by leveraging BERT as an external resource, we report new state-of-the-art $F_1$ scores of 93.47 on CoNLL03 and 97.30 on CoNLL2000.

In the future, we would like to extend GCDT to other analogous sequence labeling tasks and explore its effectiveness on other languages.

\section*{Acknowledgments}
Liu, Xu, and Chen are supported by the National Nature Science Foundation of China (Contract 61370130, 61473294 and 61502149), and Beijing Natural Science Foundation under Grant No. 4172047, and the Fundamental Research Funds for the Central Universities (2015JBM033), and the International Science and Technology Cooperation Program of China under grant No. 2014DFA11350.
We sincerely thank the anonymous reviewers for their thorough reviewing and valuable suggestions.

\bibliography{acl2019}
\bibliographystyle{acl_natbib}
\end{document}